\documentclass{article}
\PassOptionsToPackage{table}{xcolor} 
\usepackage{hyperref}

\usepackage[preprint]{icml2026}
\usepackage{microtype}

\usepackage{amsmath,amsfonts,amssymb,bm}

\def\eqref#1{equation~\ref{#1}}

\def\1{\bm{1}}

\DeclareMathAlphabet{\mathsfit}{\encodingdefault}{\sfdefault}{m}{sl}
\SetMathAlphabet{\mathsfit}{bold}{\encodingdefault}{\sfdefault}{bx}{n}

\usepackage{url}

\usepackage{enumitem}
\usepackage{xspace}

\usepackage{xcolor} 
\usepackage{tcolorbox}
\usepackage{fontawesome}  
\usepackage{listings}
\tcbuselibrary{breakable}

\usepackage{graphicx} 
\usepackage{threeparttable}
\usepackage{multirow}
\usepackage{booktabs}
\usepackage{makecell}
\usepackage{pifont}

\usepackage{epigraph}
\def\eg{{\it{e.g.}}\xspace}

\def\ie{{\it{i.e.}}\xspace}

\definecolor{lightpink}{RGB}{255, 230, 240}
\definecolor{lightblue}{RGB}{230, 240, 255}
\definecolor{lightgray}{RGB}{248, 248, 248}
\definecolor{bordergray}{RGB}{200, 200, 200}

\definecolor{bestcolor}{RGB}{219, 208, 237}
\definecolor{secondcolor}{RGB}{241, 237, 248}
\definecolor{thirdcolor}{RGB}{211, 222, 190}
\definecolor{line-blue}{RGB}{243, 248, 252}
\definecolor{ForestGreen}{RGB}{34, 139, 34}

\usepackage{wrapfig}
\definecolor{lightblue}{rgb}{0.22,0.45,0.70}%
\definecolor{lightgreen}{rgb}{0.1, 0.6, 0.1}%
\usepackage{tabularx}

\setitemize{itemsep=10pt,topsep=0pt,parsep=0pt,partopsep=0pt}
\icmltitlerunning{MentalThink: Shaping Thoughts in Mental SVG World}
\begin{document}
\twocolumn[
  \icmltitle{MentalThink: Shaping Thoughts in Mental SVG World}



  \icmlsetsymbol{core}{*}

  \begin{icmlauthorlist}
    \icmlauthor{Kangheng Lin}{core,bupt}
    \icmlauthor{Jisheng Yin}{core,ucas}
    \icmlauthor{Dingming Li}{core,step}
    \icmlauthor{En Yu}{step}
    \icmlauthor{Yana Wei}{step}
    \icmlauthor{Han Zhou}{step}
    \icmlauthor{Liang Zhao}{step}
    \icmlauthor{Hongyu Zhou}{step}
    \icmlauthor{Hongbo Peng}{step}
    \icmlauthor{Jianjian Sun}{step}
    \icmlauthor{Zheng Ge}{step}
    \icmlauthor{Xiangyu Zhang}{step}
    \icmlauthor{Daxin Jiang}{step}
    \icmlauthor{Jingyu Wang}{bupt}
  \end{icmlauthorlist}

  \icmlaffiliation{step}{StepFun}
  \icmlaffiliation{ucas}{University of Chinese Academy of Sciences}
  \icmlaffiliation{bupt}{State Key Laboratory of Networking and Switching Technology, Beijing University of Posts and Telecommunications}

  \icmlcorrespondingauthor{Jingyu Wang}{wangjingyu@bupt.edu.cn}

  \icmlkeywords{spatial reasoning, multimodal large language models, visual reasoning, mental imagery, scalable vector graphics, chain-of-thought, test-time scaling}

  \vskip 0.3in
]



\printAffiliationsAndNotice{\textsuperscript{*}Core contribution.}  

\begin{abstract}
    We introduce \textbf{\textsc{MentalThink}}, a visual-symbolic reasoning paradigm that equips Multimodal LLMs (MLLMs) with an executable mechanism for ``mental'' visualization. The core of \textsc{MentalThink} is a \textit{\textbf{think-with-SVG}} pipeline, where the model learns to generate, render, and interpret \textit{scalable vector graphics} (SVG) code as an intermediate visual representation for multi-turn reasoning. By creating structured vector sketches, the model can externalize spatial hypotheses, inspect them through deterministic rendering, and reason within a constrained geometric space, effectively mimicking the human process of mental imagery. We instantiate this paradigm through a two-stage training framework, combining Supervised Fine-Tuning (SFT) for SVG syntactic alignment with multi-turn Reinforcement Learning (RL) to encourage iterative inspection, revision, and refinement of intermediate visual hypotheses. Extensive evaluations demonstrate that \textsc{MentalThink} achieves superior performance on spatial understanding and reasoning benchmarks (\eg~55.1\% on VSIBench, 76.0\% on MindCube), showing that executable vector graphics provide a verifiable visual workspace for dynamic perspective taking, visual reflection, and compositional scene construction.
\end{abstract}

\begin{figure*}[t]
\centering
\includegraphics[width=1.0\linewidth]{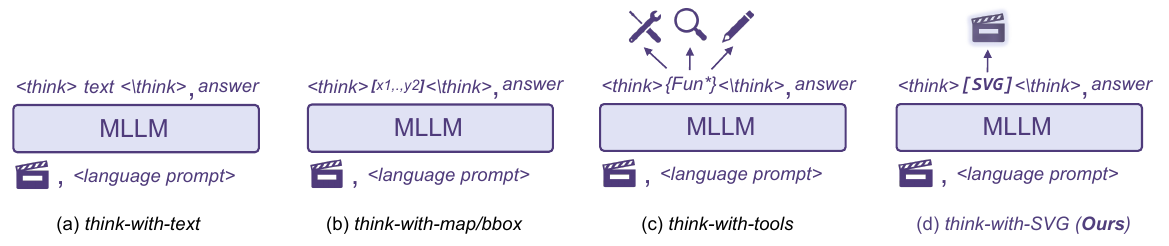}
\caption{\textbf{Conceptual comparison} of different thinking paradigms.}
\label{fig:thinking_paradigm}
\vspace{-2mm}
\end{figure*}
\section{Introduction}

\setlength{\epigraphwidth}{0.95\columnwidth}
\renewcommand{\epigraphflush}{center}
\renewcommand{\textflush}{flushepinormal}
\renewcommand{\epigraphsize}{\footnotesize}
\epigraph{\textcolor{black}{``The soul never thinks without a mental image.''}}
{\textcolor{black}{\textit{Aristotle De Anima}}}

The human capacity for reasoning extends far beyond the manipulation of linguistic symbols~\citep{dinneen1995general}. Cognitive processes often involve \textit{mental imagery}~\citep{kosslyn1980image}, where individuals construct and manipulate mental models~\citep{johnson1983mental} to understand complex relationships, plan actions, and solve problems~\citep{shepard1971mental}. This cognitive ability serves as a dynamic scratchpad for visualization, allowing us to simulate scenarios, shift perspectives, and ground abstract concepts in a \textit{quasi-physical} space. This form of visual thinking is a critical aspect of human cognition and represents a promising frontier for artificial intelligence (AI). Recent advances in large language models (LLMs) have demonstrated strong mental reasoning capabilities through long-context verbal traces that exhibit structured cognitive behaviors~\citep{liu2025prorl, wei2025open, gandhi2025cognitive}. 

Multimodal large language models (MLLMs)~\citep{guo2025seed1, hong2025glm, wei2025open} have explored diverse mechanisms to incorporate visual inputs into reasoning; however, existing paradigms inevitably incur fundamental trade-offs. Language-centric methods~\citep{wei2025open,huang2026step3} extend Chain-of-Thought (CoT)~\citep{cot} by relying on textual descriptions. Despite their flexibility, these methods often suffer from weak visual grounding: lacking structured spatial representations, textual reasoning easily drifts from geometric reality, inducing hallucinations that are semantically plausible yet visually unfaithful~\citep{liu2025more, li2026spatialevo}. Conversely, approaches leveraging explicit layout representations (e.g., 2D/3D bounding boxes, cognitive maps~\citep{yin2025spatial}) or external tools~\citep{zhang2025thymethinkimages} attempt to mitigate this issue. However, these strategies tend to either reduce rich visual scenes into sparse geometric primitives or offload visual imagination to disjoint external modules, consequently compromising the symbolic flexibility and generality inherent to language-based reasoning.

This raises a natural question: \textit{Can a reasoning pattern retain linguistic symbolic strengths while being grounded in visual structure?} We identify Scalable Vector Graphics (SVG) as a compelling solution. First, as a \textbf{structured visual language}, it offers a \textbf{complete and compositional representation} of visual layouts, encapsulating object geometry and spatial hierarchies. Consequently, generating valid vector commands requires genuine structural understanding, transcending the surface-level pattern matching often observed in pure text generation. Second, being inherently \textbf{code-based}, this representation aligns natively with the architecture of LLMs. 
Since models are extensively pretrained on both large-scale text corpora and markup or programming languages (e.g., HTML/XML), they can readily reuse existing hierarchical and compositional priors~\citep{han2025learningseeing} to manipulate visual elements expressed in code form.

To this end, we introduce \textbf{\textsc{MentalThink}}, a novel reasoning paradigm that enables multimodal LLMs to think visually through explicit and interpretable mental imagery. 
At its core, \textsc{MentalThink} adopts a \textbf{think-with-SVG} pipeline (Figure~\ref{fig:thinking_paradigm}(d)), in which the model generates SVG as an intermediate representation of its internal visual reasoning.
We realize this paradigm through a progressive two-stage training framework.
\textbf{(1)} In the first stage, we perform supervised fine-tuning (SFT) on a curated \emph{think-with-SVG} dataset, teaching the model to externalize abstract reasoning into structured SVG representations.
\textbf{(2)} In the second stage, we apply multi-turn reinforcement learning (RL) to refine the reasoning process itself, encouraging the model to iteratively inspect, revise, and correct its intermediate visual hypotheses.

Through comprehensive evaluations, \textsc{MentalThink} achieves superior performance on spatial understanding and reasoning benchmarks (\eg~55.1\% on VSIBench~\citep{yang2025thinkingspacemultimodallarge}, 76.0\% on MindCube~\citep{yin2025spatial}).
Beyond raw accuracy, our qualitative analysis reveals that \textit{think-with-SVG} elicits sophisticated spatial reasoning patterns that are difficult to replicate with pure language, which includes dynamic perspective taking, reflective visual refinement, and compositional scene construction.
These findings suggest that \textit{think-with-SVG} constitutes a promising paradigm that strengthens model cognition through explicit mental imagery, providing an efficient and verifiable abstraction for multimodal reasoning.

\section{Related Works}

\textbf{Chain-of-Thought (CoT) Reasoning}. The development of CoT prompting marked a significant milestone in eliciting complex reasoning from LLMs~\citep{cot}. By instructing models to generate a series of intermediate reasoning steps, \ie, thinking process, CoT transforms intractable problems into a sequence of manageable sub-problems. This decomposition serves as a cognitive bridge, allowing models to maintain logical coherence over long horizons and reducing the likelihood of error propagation, thus dramatically improving performance on tasks requiring complex reasoning. Such CoT processes are externalized or internalized via SFT~\citep{yue2023mammoth, yu2023metamath} and further scaled up via RL~\citep{o1, r1}, constructing a linear transformation from test-time compute to performance.

A natural evolution of CoT is its extension into the multimodal domain. Pioneering works in multimodal CoT~\citep{chatspot, merlin, zhang2023multimodal, wei2025open, meng2025mm, xiaomi2025mimo} demonstrate that MLLMs can achieve significantly better results on complex multimodal reasoning tasks when prompted or self-explored to generate a rationale that interleaves textual explanations with visual features.
However, these methods largely remain \textbf{language-centric}, where the reasoning process relies on textual descriptions of visual content. Such approaches often suffer from a lack of \textbf{visual grounding}: without a structured spatial representation, the model's textual reasoning can easily drift from the actual geometric reality of the image, leading to hallucinations that are semantically plausible but visually incorrect~\citep{liu2025more}.
Furthermore, unlike code-based reasoning which can be executed and verified, pure text lacks a mechanism for rigorous self-validation.

\textbf{Think-with-Image}. To address the limitations of language-centric reasoning, several recent methods~\citep{hu2024visual, wu2024v, o3} have emerged to ``\textit{think-with-image}''. One strategy leverages SFT or RL technology to enable models to call external tools for image manipulation, \eg, zoom-in/out~\citep{sarch2025grounded, zheng2025deepeyes} or generating auxiliary visuals~\citep{chern2025thinking, qin2025uni}, as an intermediate reasoning step.
However, the utility of this approach is often constrained by the rigid API of external tools, which breaks the end-to-end differentiability of the reasoning chain.
An alternative strategy focuses on internal visual representations, such as generating \textbf{cognitive maps}~\citep{yin2025spatial} or \textbf{bounding box layouts}~\citep{li2025imagine, yang2025machine} to explicitly model spatial relationships.
While these methods improve localization, they often oversimplify complex visual details into sparse primitives (e.g., boxes), losing the fine-grained shape information required for detailed reasoning.
In this work, we propose a novel ``\textit{think-with-SVG}'' pipeline. By generating a structured, machine-readable format that is inherently visual yet compositionally precise, we bridge the gap between abstract symbolic reasoning and concrete visual grounding without relying on black-box external tools.

\textbf{SVG in LLM \& MLLM}.
Traditionally, Scalable Vector Graphics (SVG) in the context of LLMs has been utilized primarily as a \textbf{generation target}~\cite{xing2024svgdreamer} or a \textbf{benchmark}~\cite{yang2025omnisvg} for evaluating spatial understanding.
Recent works have explored utilizing LLMs to generate SVG code for creating icons, charts, and artistic designs~\citep{rodriguez2025starvector, xing2024svgdreamer}, or used SVG-rendering tasks to assess the code-generation capabilities of models regarding spatial coordinates~\citep{wu2024v}.
However, these approaches treat SVG as the final output, the ``result'' of the thought process, rather than the medium of thought itself.
To the best of our knowledge, \textsc{MentalThink} is the first to leverage SVG as an \textit{intermediate reasoning modality}, enabling the model to construct, inspect, and refine a ``mental sketch'' to guide its decision-making process for complex multimodal tasks.

\begin{figure*}[t]
\centering
\includegraphics[width=0.98\linewidth]{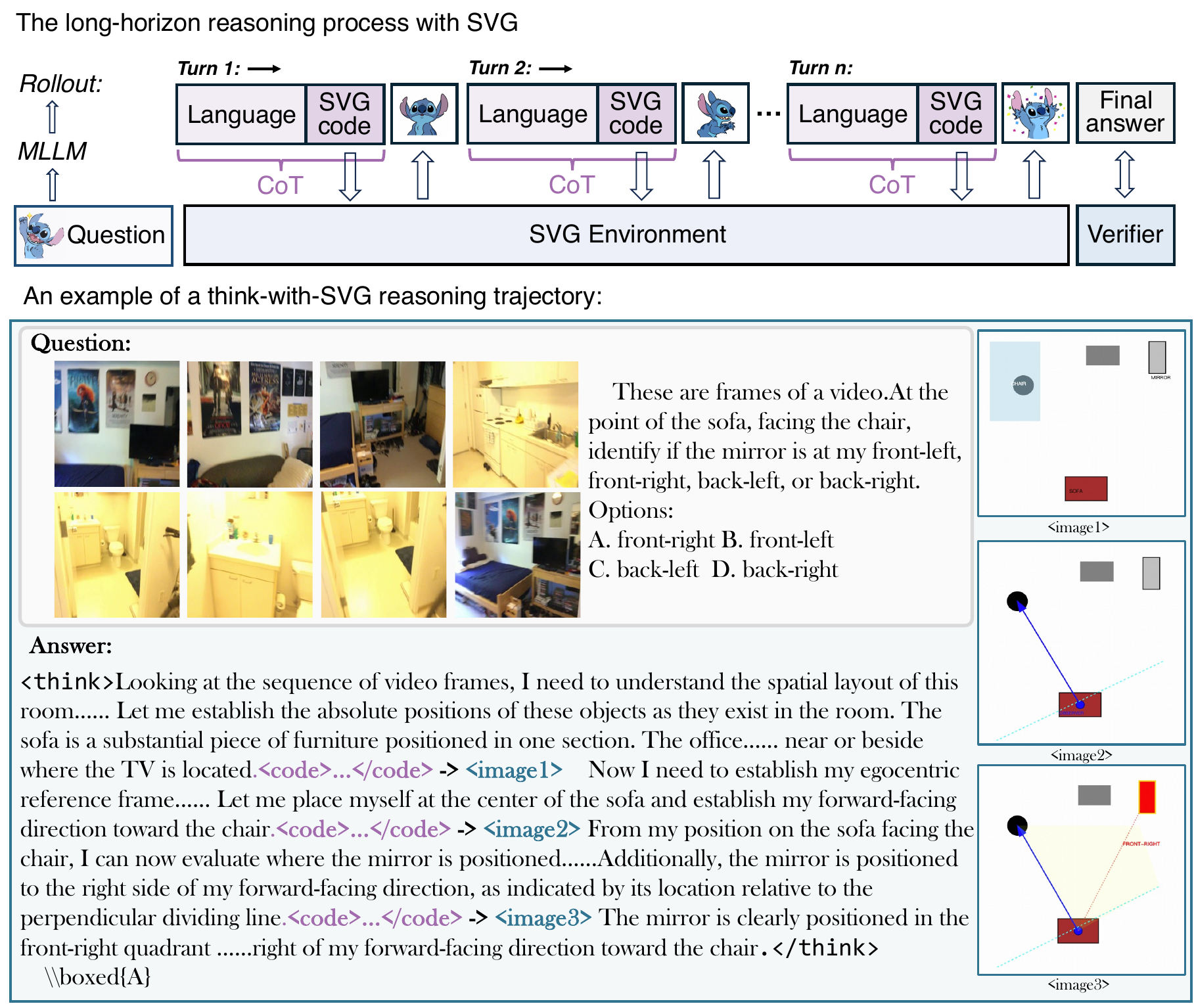}
\vspace{-1mm}
\caption{\textbf{Think-with-SVG Pipeline}. MentalThink trains MLLMs to natively use SVG code to render ``mental images'' during multi-turn thinking process with SVG environment, enabling more advanced and effective reasoning for complex problem-solving.}
\label{fig:think_with_svg}
\vspace{-4mm}
\end{figure*}

\section{Think-with-SVG Reasoning Framework}

\subsection{Preliminaries: Multimodal CoT}
Building upon recent theoretical foundations~\citep{zhang2025does, chen2025towards,gan2025cot}, we begin by establishing formal definitions for Chain-of-Thought reasoning~\citep{cot} paradigm, which structures problem-solving as a sequence of intermediate, human-readable steps.

\textbf{Definition 3.1} (Chain-of-Thought State)\label{def 3.1}. Given a query $q \in \mathcal{Q}$ and a sequence of intermediate reasoning steps $\xi = \{\xi_1, \xi_2, \ldots, \xi_t\}$, we define a CoT state as the tuple $s = (q, \xi)$, where each $\xi_i \in \mathcal{L}$ represents a semantic unit advancing toward the solution in the linguistic space $\mathcal{L}$. 

\textbf{Definition 3.2} (Reasoning Trajectory)\label{def 3.2}. A complete reasoning trajectory $\tau$ of length $T$ is a sequence $\tau = \{s_0, s_1, \ldots, s_T\}$ where $s_0 = (q, \emptyset)$ denotes the initial state and $s_T = (q, \xi, a)$ includes the final answer $a \in \mathcal{A}$. For multimodal reasoning, we extend these definitions:

\textbf{Definition 3.3} (Multimodal Chain-of-Thought)\label{def 3.3}. Given a multimodal query $(q, \mathbf{v})$ where $\mathbf{v} = \{v_1, \ldots, v_n\} \subset \mathcal{V}$ represents visual inputs, a multimodal CoT state is:
\begin{equation}
s^{\text{mm}} = (q, \mathbf{v}, \xi^{\text{mm}})
\end{equation}
where $\xi^{\text{mm}} = \{\xi_1^{\text{mm}}, \ldots, \xi_t^{\text{mm}}\}$ and each $\xi_i^{\text{mm}} \in \mathcal{L} \times \mathcal{V}$ can reference or incorporate visual information. While powerful, existing multimodal CoT process~\citep{hong2025glm, yang2025kwai, wei2025open, zheng2025deepeyes} remains fundamentally linguistic. The model verbalizes its interpretation of visual data rather than reasoning within a visual space.

\subsection{Think-with-SVG: A Structured Visual Reasoning Pipeline}

The \textit{think-with-SVG} pipeline reframes the reasoning process by replacing or augmenting purely linguistic thought steps with a structured, visual-symbolic representation. We now formalize this paradigm, establishing SVG as a structured bridge between symbolic and visual reasoning.

\textbf{Definition 3.4} (Structured Visual Language)\label{def 3.4}. We define SVG code space $\mathcal{S}$ as a structured visual language with the following properties:
\begin{equation}
\mathcal{S} = \{\sigma : \sigma = \langle E, A, H \rangle\}
\end{equation}
where $E = \{e_1, e_2, \ldots, e_n\}$ is a set of primitive elements (e.g., \texttt{<rect>}, \texttt{<circle>}, \texttt{<path>}). $A = \{a_{ij}\}$ represents attributes for each element $e_i$ (e.g., position $(x, y)$, dimensions $(w, h)$, style properties). $H$ denotes the hierarchical structure through grouping (\texttt{<g>}) and transformations. Unlike natural language descriptions, SVG enforces \textit{syntactic structure} and \textit{semantic precision}:
{\small
\begin{equation}
\sigma_{\text{valid}} \in \mathcal{S} \iff \text{Grammar}(\sigma) = \text{True} \land \text{Renderable}(\sigma) = \text{True}
\end{equation}}

This structured nature provides three key advantages: (\textbf{\textit{i}}) \textbf{Deterministic mapping} from code to visual output, (\textbf{\textit{ii}}) \textbf{Compositional semantics} where complex scenes are built from primitive elements, and (\textbf{\textit{iii}}) \textbf{Parametric control} over visual properties through explicit attributes.

\textbf{Definition 3.5} (SVG-Augmented Reasoning State)\label{def 3.5}. An SVG-augmented reasoning state extends multimodal CoT as:
\begin{equation}
s^{\text{svg}} = (q, \mathbf{v}, \xi^{\text{svg}})
\end{equation}
where $\xi^{\text{svg}} = \{(\ell_1, \sigma_1), (\ell_2, \sigma_2), \ldots, (\ell_T, \sigma_T)\}$ with $\ell_i \in \mathcal{L}$ representing linguistic reasoning and $\sigma_i \in \mathcal{S} \cup \{\emptyset\}$ representing optional SVG code. Through SVG-augmented reasoning, the model learns to adaptively generate SVG code when it needs to apply \textit{mental imagery} during thinking.

\textbf{Definition 3.6} (SVG Rendering Environment)\label{def 3.6}. The rendering process from SVG code to raster image involves a deterministic transformation:
\begin{equation}
\mathcal{R}: \mathcal{S} \rightarrow \mathcal{V}, \quad \mathcal{R}(\sigma) = \mathbf{I}_\sigma \in \mathbb{R}^{H \times W \times C}
\end{equation}

The rendering pipeline $\mathcal{R}$ consists of three sequential transformations:
\begin{equation}
\sigma \xrightarrow{\text{Parse}} \mathcal{T}_{\text{DOM}} \xrightarrow{\text{Layout}} \mathcal{G} = \{g_1, \ldots, g_m\} \xrightarrow{\text{Rasterize}} \mathbf{I}_\sigma
\label{eq:svg_rendering}
\end{equation}
where $\mathcal{T}_{\text{DOM}}$ is the document object model tree, and $\mathcal{G}$ represents geometric primitives in the coordinate space and graphic scene. Through this SVG environment, the model can iteratively generate, render, and reason about visual representations, creating a feedback loop between symbolic manipulation and visual interpretation. Figure~\ref{fig:think_with_svg} gives a detailed example showing that \textsc{MentalThink} first \textbf{establishes the absolute spatial layout}, then \textbf{defines the egocentric reference frame}, followed by \textbf{evaluating the target position} to get the final answer.

This structured visual language enables MLLMs to perform ``mental rendering'' – constructing and manipulating visual representations through code generation, thereby achieving native visual reasoning without external tools or specialized visual encoders.

\subsection{MentalThink Training Recipe}
\label{training}

\textbf{Overview.} We formulate the \textsc{MentalThink} process as optimizing a multimodal policy $\pi_\theta$ to generate optimal reasoning trajectories $\tau$. Unlike standard linguistic policies, $\pi_\theta$ operates over the augmented state space defined in Definition 3.5, where the model must learn to interleave symbolic reasoning $\ell$ with structured visual codes $\sigma \in \mathcal{S}$. We optimize $\pi_\theta$ through a progressive two-stage framework.

\textbf{Stage 1: Syntactic Alignment via Cold Start.}
To initialize the policy $\pi_\theta$ with the capability to generate valid states $s^{\text{svg}}$, we perform supervised fine-tuning (SFT).
Given a dataset of expert trajectories $\mathcal{D}_{\text{SFT}} = \{\tau^*\}$, where each step fits the definition $\xi_i = (\ell_i, \sigma_i)$, we minimize the negative log-likelihood:
\begin{equation}
\mathcal{L}_{\text{SFT}}(\theta) = - \mathbb{E}_{\tau \sim \mathcal{D}_{\text{SFT}}} \sum_{t=0}^{T} \log \pi_\theta(\xi_t^{\text{svg}} \mid s_{t}^{\text{svg}})
\end{equation}
This stage ensures the model aligns with the structural constraints of $\mathcal{S}$ (Definition 3.4), learning to generate syntactically valid SVG codes before engaging in complex reasoning. In addition, we also synthesize a small amount of \textit{think-with-SVG} cold-start data for thinking pattern warm-up. More details are shown in Appendix~\ref{sec:appendix:data-pipeline}.

\textbf{Stage 2: Strategic Reasoning via Multi-Turn RL.}
SFT provides the form, but not the reasoning dynamics. To enable self-correction, we optimize $\pi_\theta$ using Group Relative Policy Optimization (GRPO)~\citep{shao2024deepseekmath, yu2025perceptionr1pioneeringperceptionpolicy} on a reasoning corpus $\mathcal{D}_{\text{RL}}$. The optimization objective is to maximize the expected return $J(\theta) = \mathbb{E}_{\tau \sim \pi_\theta}[R(\tau)]$, where the generation process is treated as an interaction with the SVG Rendering Environment (Definition 3.6).

\vspace{1mm}
\noindent \textbf{Interactive Reasoning Loop.}
We model the multi-turn reasoning as a state transition process $s_t \to s_{t+1}$ within the SVG rendering environment (Definition 3.6):

\begin{itemize}
[leftmargin=4mm]
\setlength{\itemsep}{2pt}
    \item \textbf{Hypothesis Generation (Policy Rollout):} At step $t$, conditioned on the current history $s_t$, the policy samples an action $a_t = (\ell_t, \sigma_t) \sim \pi_\theta(\cdot | s_t)$.
    
    \item \textbf{Deterministic Visual Feedback (SVG Rendering):} If the action includes non-empty SVG code ($\sigma_t \neq \emptyset$), the environment triggers the rendering function defined in Eq.~\ref{eq:svg_rendering}:
    \begin{equation}
    \mathbf{I}_t = \mathcal{R}(\sigma_t)
    \end{equation}
    This step effectively converts the model's symbolic hypothesis $\sigma_t$ into pixel-space observation $\mathbf{I}_t$.
    
    \item \textbf{State Transition (Context Update):} The rendered image is appended to the visual context, updating the state:
    \begin{equation}
    s_{t+1} = s_t \oplus (\ell_t, \sigma_t, \mathbf{I}_t)
    \end{equation}
    This explicitly closes the loop, allowing $\pi_\theta$ to condition its next reasoning step $\xi_{t+1}$ on the visual verification of its previous thought.
    
    \item \textbf{Termination:} The trajectory terminates when the model outputs a stop token, reaches $T_{\max}$, or triggers a hard syntax failure ($\sigma_t \notin \mathcal{S}$).
\end{itemize}

\vspace{1mm}

\noindent \textbf{Reward Formulation.}
The reward $R(\tau)$ is designed to enforce the structural integrity defined in Definition 3.4 and the final correctness:
\begin{equation}
    r_t = \lambda_{\mathrm{fmt}}\, r_{\mathrm{format}}(\sigma_t) + \lambda_{\mathrm{ans}}\, r_{\mathrm{answer}}(a_T) \cdot \mathbb{I}(t=T)
    \label{eq:reward}
\end{equation}

\begin{itemize}
[leftmargin=4mm]
\setlength{\itemsep}{2pt}

\item \textbf{Format Reward ($r_{\mathrm{format}}$):} A dense reward that validates if $\sigma_t$ satisfies the grammar of $\mathcal{S}$. Specifically, $r_{\mathrm{format}} = 1$ iff $\text{Renderable}(\sigma_t) = \text{True}$, ensuring the ``mental image'' is valid.

\item \textbf{Answer Reward ($r_{\mathrm{answer}}$):} A sparse reward awarded only at the final step $T$, measuring if the predicted answer matches the ground truth.

\end{itemize}

\section{Experiments}
\label{sec:experiments}

\begin{table*}[ht!]
    \centering
    \caption{\textbf{Evaluation on spatial understanding and reasoning benchmarks}.\textbf{MindCube*} denotes MindCube-Tiny. \textbf{Improve} indicates gains over the Qwen2.5-VL-7B backbone. Dark/light purple highlights the best/second-best results in Proprietary and Open-source categories, respectively.}
    \label{tab:main_results}

    \setlength{\tabcolsep}{2.5pt}
    \resizebox{0.76\linewidth}{!}{

    \begin{tabular}{lcccccc}
    \toprule
    \textbf{Models} &
    \textbf{VSIBench} &
    \textbf{MMSI} &
    \textbf{OmniSpatial} &
    \textbf{MindCube$^*$} &
    \textbf{SpatialViz} &
    \textbf{ViewSpatial} \\

    \midrule
    \rowcolor{line-blue}\textit{\textbf{Proprietary Models}} & & & & & & \\

    Seed-1.6-2025-06-15 &  49.9 & \cellcolor{secondcolor}{38.3} & \cellcolor{bestcolor}{\textbf{54.5}} & 48.7 & \cellcolor{bestcolor}{\textbf{34.9}} & 43.8 \\

    Gemini-2.5-pro-2025-06 & \cellcolor{secondcolor}{53.5} & 38.0 & 43.0 & \cellcolor{bestcolor}{\textbf{57.6}} & 15.9 & \cellcolor{bestcolor}{\textbf{46.0}} \\

    GPT-5-nano-2025-08-07 & 38.9 & 22.5 & 42.3 & 19.0 & 4.5 & 38.5 \\

    GPT-5-mini-2025-08-07 & 48.2 & 34.3 & \cellcolor{secondcolor}{52.0} & 50.0 & \cellcolor{secondcolor}{22.5} & 42.4 \\

    GPT-5-2025-08-07 & \cellcolor{bestcolor}{\textbf{55.0}} & \cellcolor{bestcolor}{\textbf{41.8}} & 49.5 & \cellcolor{secondcolor}{56.3} & 7.2 & \cellcolor{secondcolor}{45.5} \\
    \midrule

    \rowcolor{line-blue}\textit{\textbf{Open-source Models}} & & & & & & \\

    Qwen2.5-VL-3B & 27.0 & \cellcolor{secondcolor}{28.6} & 40.3 & 34.7 & 30.2 & 31.9 \\
    \textbf{Qwen2.5-VL-7B} & 31.0 & 27.8 & 39.2 & 36.0 & 26.8 & 36.8 \\
    Qwen2.5-VL-72B & 34.7 & \cellcolor{bestcolor}{\textbf{30.7}} & \cellcolor{secondcolor}{47.9} & \cellcolor{secondcolor}{37.0} & \cellcolor{bestcolor}{\textbf{35.0}} & \cellcolor{bestcolor}{\textbf{38.9}} \\

    LLaVA-Next-Video-7B & 35.6 & - & 28.5 & - & 20.9 & 30.6 \\
    LLaVA-Next-Video-72B & \cellcolor{secondcolor}{40.9} & - & - & - & - & - \\

    LLaVA-OneVision-7B & 32.4 & 24.5 & 35.7 & - & 24.7 & 27.5 \\
    LLaVA-OneVision-72B & 40.2 & 28.4 & 45.7 & - & - & - \\

    InternVL3-8B & \cellcolor{bestcolor}{\textbf{42.1}} & 28.0 & 41.6 & \cellcolor{bestcolor}{\textbf{41.5}} & 30.3 & \cellcolor{secondcolor}{38.6} \\
    InternVL3-78B & - & 28.5 & \cellcolor{bestcolor}{\textbf{49.3}} & - & \cellcolor{secondcolor}{32.3} & - \\

    \midrule

    \rowcolor{line-blue}\textit{\textbf{Spatial-Specific Models }} & & & & & & \\

    MindCube-3B-RawQA-SFT & 17.2 & 1.7 & 24.5 & 51.7 & 26.5 & 24.1 \\

    SpatialLadder-3B & 45.7 & 27.4& 41.9 & 43.4 & 28.4 & 44.2 \\

    Spatial-MLLM-4B & 47.3 & 26.1 & 38.0 & 33.4 & - & 34.6 \\

    SpaceR-SFT-7B & 41.6 & 27.4& 41.0 & 38.0 & 29.2 & 35.8 \\

    VILASR-7B & 44.6 & \cellcolor{bestcolor}{\textbf{30.2}} & 19.2 & 35.1 & 31.3 & 35.7 \\

    \textbf{\textsc{MentalThink}-SFT} & \cellcolor{secondcolor}{53.9} & 24.5 & \cellcolor{bestcolor}{\textbf{44.9}} & \cellcolor{secondcolor}{74.2}& \cellcolor{secondcolor}{34.2} & \cellcolor{secondcolor}{62.3} \\

    \textbf{\textsc{MentalThink}-RL} & \cellcolor{bestcolor}{\textbf{55.1}}& \cellcolor{secondcolor}{28.0}& \cellcolor{secondcolor}{43.4}& \cellcolor{bestcolor}{\textbf{76.0}}& \cellcolor{bestcolor}{\textbf{38.7}}& \cellcolor{bestcolor}{\textbf{62.5}} \\

    \midrule

    \rowcolor{gray!6}
    \textit{Improve} $\uparrow$ & \textcolor{lightgreen}{+24.1}& \textcolor{lightgreen}{+0.2}& \textcolor{lightgreen}{+4.2}& \textcolor{lightgreen}{+40.0}& \textcolor{lightgreen}{+11.9}& \textcolor{lightgreen}{+25.7}
    \\

    \bottomrule
    \end{tabular}
    }
    \end{table*}

\subsection{Experimental Setup}
\label{sec:setup}

\noindent \textbf{Implementation Setup and Baselines.} We implement \textsc{MentalThink} atop the \textbf{Qwen2.5-VL-7B} backbone~\citep{qwenvl2p5}. This choice allows us to strictly \textbf{isolate} the empirical gains attributable to the \textit{think-with-SVG} paradigm from those potentially stemming from mere model scaling. 
We comprehensively compare \textsc{MentalThink} against three categories of competitive baselines:
\begin{itemize}[leftmargin=4mm, itemsep=2pt]
    \item \textbf{Proprietary Models:} Seed-1.6~\citep{guo2025seed1}, Gemini-2.5-pro~\citep{team2023gemini}, and the GPT-5 series (nano, mini, full)~\citep{singh2025openaigpt5card}.
    \item \textbf{Open-source General MLLMs:} Qwen2.5-VL~\citep{qwenvl2p5} (3B/7B/72B), LLaVA-Next-Video~\citep{llavanext-video} (7B/72B), LLaVA-OneVision~\citep{li2024llavaonevisioneasyvisualtask} (7B/72B), and InternVL3~\citep{zhu2025internvl3} (8B/78B).
    \item \textbf{Spatial-Specific Models:} MindCube-3B-RawQA-SFT, SpatialLadder-3B~\citep{li2025spatialladderprogressivetrainingspatial}, Spatial-MLLM-4B~\citep{wu2025spatialmllmboostingmllmcapabilities}, SpaceR-SFT-7B~\citep{ouyang2025spacerreinforcingmllmsvideo}, and VILASR-7B~\citep{wu2025reinforcingspatialreasoningvisionlanguage}.
\end{itemize}

\noindent \textbf{Data Synthesis and Composition.} We utilize the multimodal dataset curated in Section~\ref{training}, specifically tailored for our two-stage training strategy. The supervised fine-tuning (SFT) corpus consists of approximately \textbf{200k} samples, categorized into three functional pillars: 
(1) \textbf{Visual-Syntactic Alignment (50k)} to bridge pixel-level perception with vector-based descriptions;
(2) \textbf{Visual Thought Externalization (50k)} featuring explicit SVG-mediated reasoning chains; 
(3) \textbf{Fundamental Spatial Perception (100k)} sampled from Euclid~\citep{lian2025euclidsgiftenhancingspatial}, DrivingVQA~\citep{corbiere2025retrievalbasedinterleavedvisualchainofthought}, SAT~\citep{sat}, SpaceVista~\citep{spacevista}, VLM-3R~\citep{fan2025vlm3rvisionlanguagemodelsaugmented}, and Spatial-SSRL~\citep{liu2025spatialssrlenhancingspatialunderstanding}.  

Our SVG reasoning and reconstruction data are constructed through a teacher-student pipeline and filtered with SVG structural validity and basic visual integrity checks. Detailed prompts, filtering rules, and dataset construction procedures are provided in Appendix~\ref{sec:appendix:data-pipeline}. For the subsequent RL stage, we use prompts from the \emph{training} splits of \textbf{VSIBench} and \textbf{MindCube}.

\noindent \textbf{Training Protocols.} All experiments are conducted on a cluster of 32 NVIDIA H800 GPUs. We employ Group Relative Policy Optimization (GRPO) driven by the hybrid reward function defined in Eq.~\ref{eq:reward} (with coefficients $\lambda_{fmt}=0.25$ and $\lambda_{ans}=0.75$). The maximum reasoning horizon for iterative refinement is capped at $T_{max}=5$. To balance convergence and stability, we adopt a progressive training schedule:
\begin{itemize}[leftmargin=4mm, itemsep=2pt]
    \item \textbf{SFT Stage:} The model is trained with a learning rate of $5\times10^{-5}$ and a global batch size of $256$.
    \item \textbf{RL Stage:} To ensure policy stability, we transition to a more conservative learning rate of $2\times10^{-6}$ and a global batch size of $64$.
\end{itemize}

\subsection{Main Results}
\label{sec:main_results}

We evaluate \textsc{MentalThink} across a comprehensive suite of spatial understanding and reasoning benchmarks, including \textbf{VSIBench}, \textbf{MMSI-Bench}~\citep{yang2025mmsi}, \textbf{OmniSpatial}~\citep{jia2025omnispatial}, \textbf{MindCube}, \textbf{SpatialViz}~\citep{wang2025spatialvizbenchmllmbenchmarkspatial}, and \textbf{ViewSpatial}~\citep{li2025viewspatialbenchevaluatingmultiperspectivespatial}. Table~\ref{tab:main_results} summarizes the performance trajectory from the baseline to the SFT intermediate, and finally to the RL-optimized model.

\noindent \textbf{Superior Spatial Performance.} As evidenced in Table~\ref{tab:main_results}, \textsc{MentalThink} demonstrates \textbf{substantial improvements} on geometry-intensive tasks. Most notably, on \textbf{MindCube} that demands complex 3D mental rotation and spatial manipulation, our method improves accuracy from 36.0\% to \textbf{76.0\%}. This remarkable gain of \textbf{40 points} \textbf{validates} our core hypothesis: explicitly externalizing thought processes into SVG code enables the model to resolve spatial relationships that \textbf{remain intractable} for purely linguistic reasoning. Consistent gains are also observed on \textbf{VSIBench} (+24.1\%) and \textbf{ViewSpatial} (+25.7\%), demonstrating \textsc{MentalThink}'s powerful spatial perception and reasoning abilities.

\noindent \textbf{Synergy of SFT and RL.} The performance comparison in Table~\ref{tab:main_results} between SFT and RL stages highlights the necessity of the two-stage pipeline. While SFT establishes the capability to generate valid SVG (``drawing mental image''), the RL stage is crucial for \textbf{mastering complex reasoning logic} (``thinking in mental image''). We observe that RL boosts performance on the most challenging logical tasks (e.g., MindCube $+1.8\%$, SpatialViz $+4.5\%$). 

\subsection{Ablation Studies}
\label{sec:ablation}

In this section, we aim to conduct comprehensive ablation studies on the proposed \textsc{MentalThink} framework, covering data design and training paradigms. From the results shown in Table~\ref{tab:ablation_overall}, we can observe:

\textbf{SVG data enhances spatial perception.} Comparing rows \#1 and \#2 in Table~\ref{tab:ablation_overall}, incorporating SVG-Reasoning data yields a foundational performance gain of 13.4\% in average score, rising from 31.0\% to 44.4\%. By framing spatial reasoning as structured SVG reconstruction, the model develops a more granular perception of layouts. This effect is most prominent in the \textbf{Appearance Order} metric, which nearly doubles from 28.2 to 56.0, and \textbf{Object Size}, which surges from 37.2 to 55.9. These results indicate that the SVG modality effectively serves as a structured inductive bias, bridging the gap between raw visual inputs and precise spatial grounding. 

\begin{table}[htbp]
\centering
\small
\caption{
\textbf{Ablation study of our proposed components on VSIBench.}
The results demonstrate that SVG-reasoning data, general spatial perception data, and RL training each provide complementary performance gains (\ding{51}/\ding{55} denote whether each component is enabled). The integrated model (last row) achieves the best performance across all metrics.
}
\label{tab:ablation_overall}
\setlength\tabcolsep{3pt} 
\resizebox{\columnwidth}{!}{
\begin{tabular}{ccc|c|ccccc}
\toprule
\multirow{2}{*}{\makecell[c]{\textbf{SVG-Reas.}\\\textbf{Data}}} &
\multirow{2}{*}{\makecell[c]{\textbf{Gen-Perc.}\\\textbf{Data}}} &
\multirow{2}{*}{\makecell[c]{\textbf{RL}\\\textbf{Task}}} &
\multirow{2}{*}{\raisebox{-0.5ex}{\textbf{Avg.}}} &
\multicolumn{2}{c}{\textbf{Numerical Question}} &
\multicolumn{3}{c}{\textbf{Multiple-Choice Question}} \\
\cmidrule(lr){5-6}\cmidrule(lr){7-9} &
~ & ~ &  ~ & Obj. Size & Room. Size & Rel. Dir. & Rel. Dist. & Appr. Order.\\
\midrule
\ding{55} & \ding{55} & \ding{55} & 31.0 & 37.2 & 43.2 & 36.3 & 37.3 & 28.2 \\
\ding{51} & \ding{55} & \ding{55} & 44.4 & 55.9 & 31.9 & 41.9 & 47.6 & 56.0 \\
\ding{51} & \ding{51} & \ding{55} & 53.9 & 64.2 & 58.4 & 47.1 & 55.5 & 63.8 \\
\ding{51} & \ding{51} & \ding{51} & \textbf{55.1} & \textbf{64.8} & \textbf{58.9} & \textbf{51.4} & \textbf{56.5}& \textbf{66.7} \\
\bottomrule
\end{tabular}
}

\vspace{-3mm}
\end{table}

\textbf{General data yields the complementary gains.} The addition of General Spatial Perception Data results in a substantial further improvement, raising the average performance from 44.4 to 53.9. This module provides broad spatial priors necessary to master scene-level attributes that specialized SFT might overlook. Notably, the \textbf{Room Size} metric experiences a massive recovery and gain, increasing by 26.5 points (from 31.9 to 58.4). This highlights that extensive and diverse spatial exposure is essential for robust environment modeling and overcoming the limitations of domain-specific data.

\textbf{RL optimizes complex reasoning.} The multi-turn RL stage acts as the catalyst for overcoming reasoning bottlenecks observed in SFT-only variants. While SFT models (row \#3) achieve strong results in perception, RL training further incentivizes the model to iteratively refine its internal hypotheses, pushing the average score to a peak of 55.1\%. This improvement is consistent across challenging logical tasks, with \textbf{Relative Direction} increasing from 47.1 to 51.4 and \textbf{Appearance Order} reaching 66.7. This framework provides the "System 2" slow thinking capability, allowing the model to convert additional inference-time computation into higher reasoning precision through verified visual simulation.

Ultimately, the full configuration of \textsc{MentalThink} achieves peak performance across all metrics. This integrated pipeline successfully bridges the gap between abstract reasoning and concrete grounding by synthesizing high-quality spatial data with iterative reinforcement learning.

\section{Analysis and Discussion}

\subsection{Empirical Analysis of Think-with-SVG}

\begin{figure*}[t!]
\centering
\vspace{-3mm}
\includegraphics[width=0.80\linewidth]{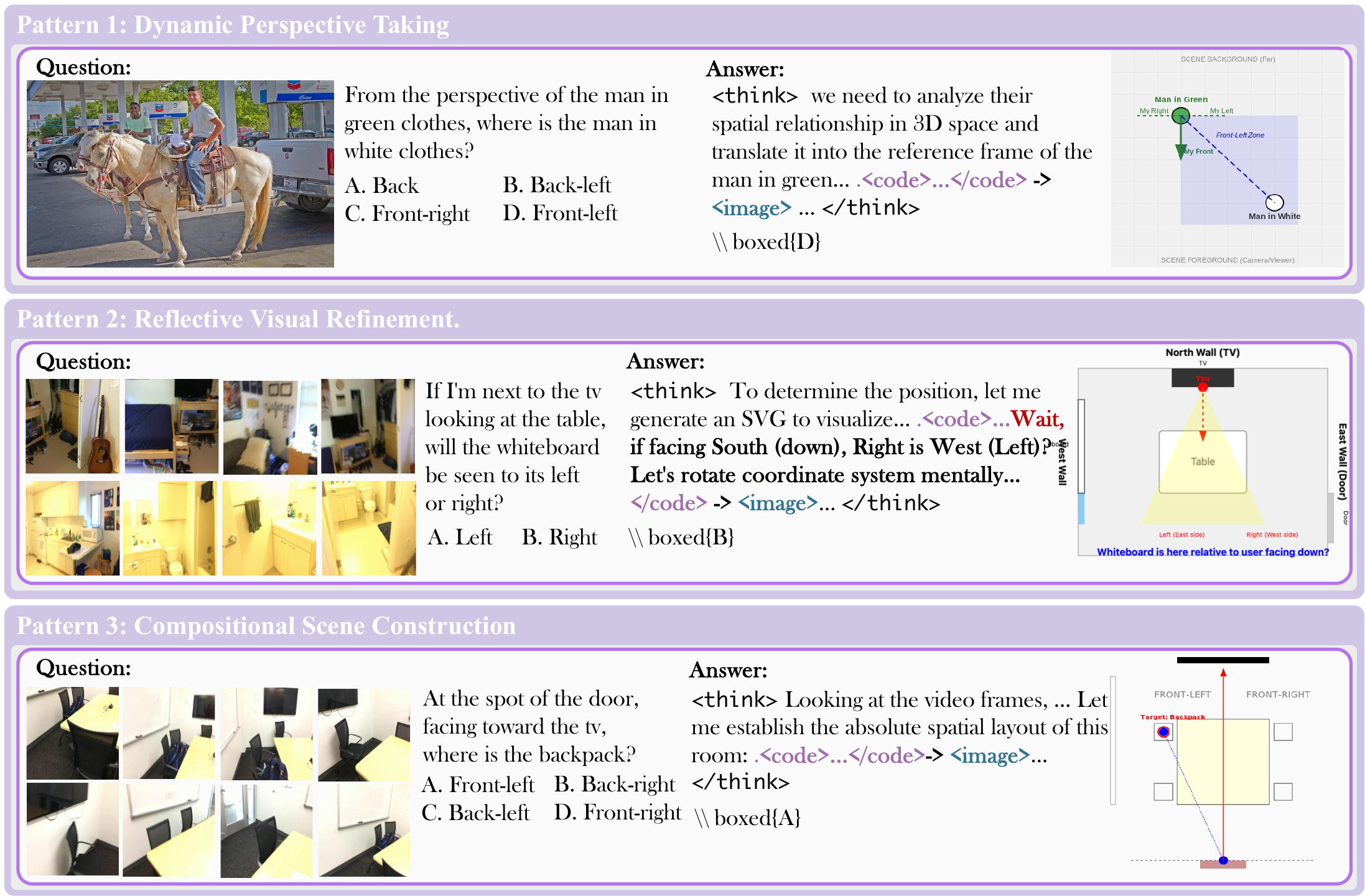}
\vspace{-2mm}
\caption{\textbf{Examples of Spatial Reasoning Patterns.} \textsc{MentalThink} demonstrates dynamic perspective taking, reflective visual refinement, and compositional scene construction.}
\label{fig:good_patterns}
\vspace{-4mm}
\end{figure*}

\textbf{Analysis of different thinking paradigms.}
We evaluate three thinking paradigms—Think-with-Text, Think-with-BBox, and Think-with-SVG—under a controlled setting where the same teacher model, task inputs, and supervision signals are used, differing only in the form of intermediate reasoning representation.

\begin{table}[htbp]
\centering
\small
\caption{
\textbf{Performance comparison between different thinking paradigms on MindCube and VSIBench.}
}
\label{tab:thinking_paradigms}
\setlength\tabcolsep{6pt}
\begin{tabular}{l|c|c}
\toprule
\textbf{Thinking Paradigm} 
& \textbf{MindCube}
& \textbf{VSIBench} \\
\midrule
\textit{Think-with-Text }
& 42.5
& 24.5 \\

\textit{Think-with-BBox}
& 41.1 
& 17.4 \\

\textit{Think-with-SVG} (ours) 
& \textbf{57.8} 
& \textbf{39.4} \\
\bottomrule
\end{tabular}
\vspace{-3mm}
\end{table}

As reported in Table~\ref{tab:thinking_paradigms}, SVG-based reasoning consistently outperforms text-based and box-based reasoning on both MindCube and VSIBench, with the largest relative gains observed on VSIBench.

These results suggest that structured and executable spatial representations provide a more effective inductive bias for complex multi-step visual reasoning.

\textbf{System 1 vs. System 2 Behavior via Adaptive Reasoning.}
As shown in Table~\ref{tab:system1_system2}, we observe a systematic shift in inference behavior across task types. We quantify this computational cost via three metrics: Avg. Turns, representing the total execution cycles where a baseline LLM response counts as one turn; SVG Invoc., the rate of externalizing thought into vector code; and Avg. Tokens/Ex., the total information density per response. For perception-dominant tasks, the model exhibits System 1-style intuition, providing direct answers with only 1.19 average turns and a low 18.5\% SVG invocation rate.

\begin{table}[htbp]
\centering
\small
\caption{
\textbf{System 1 vs. System 2 style behavior across task types.} We report the average reasoning turns, SVG invocation rate, average token length per example, and final task accuracy. Notably, the model adapts its reasoning effort (reasoning turns, SVG invocation, and token usage) to task complexity while maintaining comparable final accuracy across both regimes.
}
\label{tab:system1_system2}
\setlength\tabcolsep{6pt}
\resizebox{\columnwidth}{!}{
\begin{tabular}{l|ccc|c}
\toprule
\textbf{Task Type} & \textbf{Avg. Turns} & \textbf{SVG Invoc.} & \textbf{Avg. Tokens/Ex.} & \textbf{Accuracy} \\
\midrule
Intuitive Visual Perception      & 1.19 & 18.5\% & 109.2  & 55.11 \\
Compositional Visual Reasoning & 1.64 & 63.7\% & 544.5 & 55.10 \\
\bottomrule
\end{tabular}
}
\vspace{-3mm}
\end{table}

In contrast, complex visual reasoning tasks trigger a deliberate System 2 process, characterized by a surge in SVG invocation (63.7\%) and token length (544.5 vs. 109.2). The increase to 1.64 average turns for reasoning tasks validates an adaptive allocation of effort. This test-time scaling demonstrates that \textsc{MentalThink} autonomously invokes iterative mental imagery to transform additional inference-time compute into rigorous visual grounding when faced with intricate spatial logic.

\subsection{Case Analysis: Emergent Spatial Reasoning via Think-with-SVG}

Figure~\ref{fig:good_patterns} shows how \textit{think-with-SVG} operationalizes mental imagery as an explicit and revisable spatial workspace, enabling structured manipulation of reference frames and object relations beyond surface-level visual description.

\textbf{Dynamic Perspective Taking.}
In the top case, the model anchors an egocentric coordinate frame on the man in green and projects the man in white into this frame, converting camera-centric observations into agent-centric spatial relations. Our model explicitly encodes forward and left--right axes through SVG generation during thinking, enabling reliable perspective transformation.

\textbf{Reflective Visual Refinement.}
In the middle case, the model constructs an initial room layout around the TV and table, then revises left--right relations after detecting a mismatch between assumed orientation and the user’s facing direction. This correction is realized through iterative updates to the SVG, which serves as a visual scratchpad for hypothesis validation.

\textbf{Compositional Scene Construction.}
In the bottom case, the model integrates multiple views by fixing the door as a reference point and aligning the table and backpack into a shared spatial layout. The composed SVG supports inferring the backpack’s front-left relation, which depends on global scene consistency rather than any single view.

Overall, these cases indicate that \textit{think-with-SVG} induces a structured form of visual reasoning in which spatial hypotheses are explicitly instantiated, inspected, and revised within a mental image, enabling consistent perspective alignment and global spatial coherence beyond language-only reasoning.

\section{Limitation and Conclusion}
\label{limitation_and_conclusion}

Representing the physical world effectively is a fundamental challenge for multimodal intelligence. We present evidence that Scalable Vector Graphics (SVG) serve as a structured, visual-symbolic representation bridging symbolic reasoning and visual perception. By treating SVG as an executable ``mental scratchpad,'' our \textit{think-with-SVG} pipeline enables models to ground their thoughts in precise geometric structures, and within \textsc{MentalThink} achieves state-of-the-art performance across multiple spatial reasoning benchmarks.

However, empowering the model to spontaneously activate and explore this ability remains a non-trivial challenge. High-quality chain-of-thought data interwoven with valid, meaningful SVG code is scarce in standard pre-training corpora. Consequently, relying on naturally occurring data is insufficient; sophisticated data synthesis and reinforcement strategies are required to bridge this gap. We hope this work can serve as a foundational step toward neuro-symbolic multimodal systems, inspiring future exploration into how explicit visual imagery can drive more robust and interpretable machine reasoning.

\clearpage
\section*{Impact Statements}
This paper presents work whose goal is to advance the field of machine learning. There are many potential societal consequences of our work, none of which we feel must be specifically highlighted here.

\bibliography{example_paper}
\bibliographystyle{icml2026}


\newpage
\clearpage
\appendix
\section{Appendix}
\label{sec:appendix}

\subsection{Data Pipeline}
\label{sec:appendix:data-pipeline}

\paragraph{Overview.}
We construct training data via a teacher-student pipeline, using \textbf{Gemini 3}~\citep{gemini3pro} as the teacher model to generate SVG-based reasoning traces. The data underpin the two-stage training recipe by providing both structured SVG reasoning demonstrations and reconstruction targets.

\paragraph{Data Sources.}
The 50k \textbf{Visual Thought Externalization} data is drawn from the training splits of \textbf{MindCube} (10k) and \textbf{VSIBench} (40k); from VSIBench we keep items that require strong visual reasoning.

\paragraph{Prompt Templates.}
We use two prompt templates to generate the SVG reasoning and reconstruction data.

\begin{tcolorbox}[title={\textbf{Prompt A: SVG Reasoning}}, colback=lightblue!18, colframe=lightblue!70!black, boxrule=0.6pt, arc=2pt, breakable]
\begin{lstlisting}
You are an expert in visual thinking who can solve problems through visualization. Whether facing mathematical, logical, spatial, system design, or process analysis challenges, you leverage SVG graphics as thinking tools.

## Core Principles
- SVG is not decoration, but a thinking instrument
- Graphic complexity should match problem needs (simple graphics for simple problems)
- Each modification should bring new understanding

## Problem-Solving Steps

### 1. [Problem Modeling]
Visualize core elements with SVG:
- What shapes represent entities/concepts?
- How to show relationships/connections?
- How to annotate constraints/conditions?

### 2. [Exploratory Analysis]
Deepen understanding through SVG manipulation:
- Transformation: Can rotation, scaling, or movement reveal new perspectives?
- Decomposition: Can complex problems be broken into simple parts?
- Composition: Do combinations of elements create new patterns?

### 3. [Pattern Recognition]
Annotate discovered patterns in SVG:
- Symmetry, periodicity, recursive structures
- Bottlenecks, critical paths, equilibrium points
- Hidden correlations, unexpected simplifications

### 4. [Solution Construction]
Show solution paths with SVG:
- For sequential processes, show the sequence
- For structural problems, show the architecture
- For dynamic problems, show the transformation

## Application Scenarios

**Math/Geometry**: Shape transformations, spatial relationships, function visualization
**System Design**: Architecture diagrams, data flows, state machines
**Logical Reasoning**: Venn diagrams, decision trees, causal chains
**Process Analysis**: Flowcharts, timelines, dependencies
**Data Understanding**: Distribution plots, relationship networks, hierarchies

## Response Requirements

For each SVG thinking step, explain:
1. [Design Intent] Why choose this visualization approach?
2. [Observations] What does this graphic reveal?
3. [Thinking Progress] How does this change your understanding?

Remember: Good visualization makes complexity simple, implicit explicit, and abstract concrete.
\end{lstlisting}
\end{tcolorbox}

\begin{tcolorbox}[title={\textbf{Prompt B: SVG Reconstruction}}, colback=lightpink!18, colframe=lightpink!70!black, boxrule=0.6pt, arc=2pt, breakable]
\begin{lstlisting}
You are an expert SVG code generator. Your task is to analyze the provided image and generate precise SVG code that accurately reconstructs the visual elements in the image.

Guidelines:

Carefully observe all visual elements including shapes, colors, text, gradients, patterns, and spatial relationships
Generate clean, valid SVG code that can be directly rendered in any browser
Use appropriate SVG elements (path, rect, circle, ellipse, polygon, text, etc.) based on the shapes in the image
Preserve accurate proportions, positions, and color values (use hex codes or rgb values)
Optimize the code for readability while maintaining accuracy
Include proper SVG header with viewBox and dimensions
Use groups (<g>) and semantic IDs when appropriate for organization
For complex shapes, use path elements with precise coordinates
If the image contains text, recreate it using SVG text elements with appropriate fonts
Add comments to explain complex sections if needed
Output format:

Provide complete, standalone SVG code wrapped in ```svg code blocks
Ensure the code is properly formatted and indented
The SVG should be scalable and maintain aspect ratio
Test that all paths are closed properly and colors are accurate
Your goal is to create SVG code that, when rendered, produces a faithful reproduction of the original image.
Answer only the SVG code, no other text.
\end{lstlisting}
\end{tcolorbox}

\paragraph{Sources and Licensing.}
We use the public \textbf{MindCube} and \textbf{VSIBench} datasets under their original licenses and usage terms. We do not redistribute raw data; instead, we report results on their official evaluation splits. We exclude samples that are corrupted or fail basic visual integrity checks. From the \textbf{590k} VSIBench training set, we retain a \textbf{40k} subset that requires strong visual reasoning.

\subsection{Thinking with Different Patterns}
\label{sec:appendix:thinking-patterns}
To study how different intermediate reasoning representations affect learning and generalization, we construct three controlled variants of system prompts that induce distinct \emph{thinking patterns} from the same teacher model.
All variants share identical task inputs, output requirements, and supervision signals, differing only in the form of intermediate reasoning representation.

\paragraph{Thinking Pattern Abstraction.}
A thinking pattern is the structured intermediate representation the teacher model uses to externalize its reasoning. We explore three:
(i) \textbf{Text-based thinking}, expressed in natural language;
(ii) \textbf{Box-based thinking}, grounded in spatial bounding boxes;
(iii) \textbf{SVG-based thinking}, expressed through executable vector graphics.
The Text- and Box-based variants use \textbf{Prompt C} and \textbf{Prompt D} below, respectively, while the SVG-based variant uses the \textbf{SVG Reasoning} prompt (\textbf{Prompt A}) from the Data Pipeline above.

\begin{tcolorbox}[title={\textbf{Prompt C: Text-based Reasoning}}, colback=gray!12, colframe=gray!70!black, boxrule=0.6pt, arc=2pt, breakable]
\begin{lstlisting}
You must reason using ONLY natural language.

Inside <think>:
- Describe objects, spatial relations, and reasoning steps using text only.
- Do NOT use coordinates, bounding boxes, SVG, diagrams, or symbolic layouts.
- Do NOT invent precise measurements.
- Express uncertainty explicitly when needed.

The <think> section should read like a careful human reasoning process
that interprets the visual scene verbally.

After <think>, output the final answer in \\boxed{}.
\end{lstlisting}
\end{tcolorbox}

\begin{tcolorbox}[title={\textbf{Prompt D: Bounding-box Reasoning}}, colback=lightgreen!18, colframe=lightgreen!70!black, boxrule=0.6pt, arc=2pt, breakable]
\begin{lstlisting}
You must reason using structured bounding box representations.

Inside <think>:
- Represent each relevant object using an axis-aligned bounding box.
- Use normalized coordinates in [0, 1000].
- Format each object as:
  ObjectName: bbox(x_min, y_min, x_max, y_max)
- Use bounding boxes to infer spatial relationships.
- Do NOT use SVG, curves, paths, or free-form drawings.
- Do NOT describe shapes beyond what bounding boxes can represent.

After listing bounding boxes, explicitly reason about spatial relations
(left/right/front/behind/overlapping) based on the boxes.

After <think>, output the final answer in **\\boxed{}**.
\end{lstlisting}
\end{tcolorbox}

\subsection{More Case Studies}
\label{sec:appendix:cases}

\begin{figure}[t]
\centering
\includegraphics[width=1.0\linewidth]{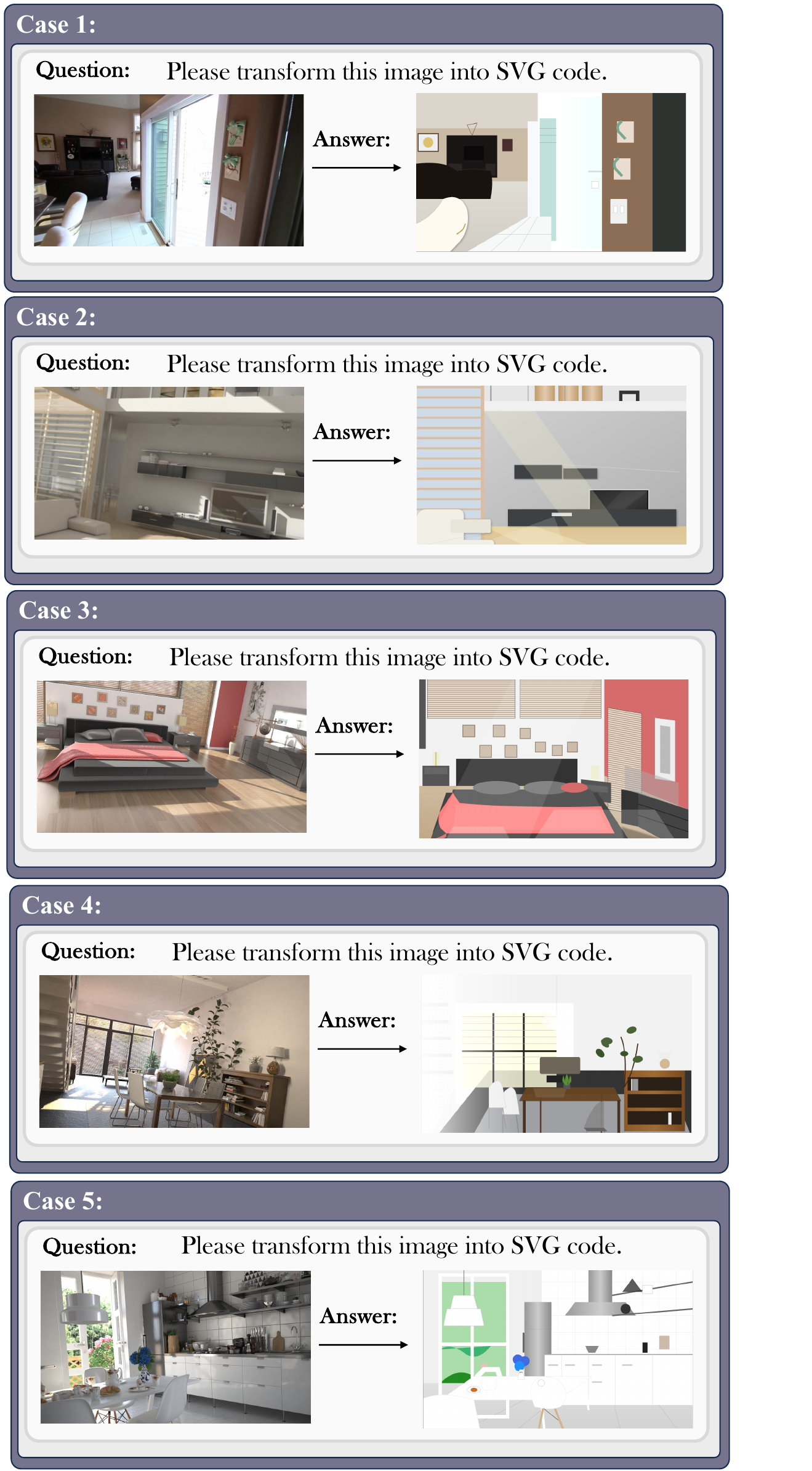}
\caption{\textbf{Additional case studies} for the SVG reconstruction capability of \textsc{MentalThink}.}
\label{fig:cases_reconstruction}
\end{figure}

\begin{figure*}[t]
\centering
\includegraphics[width=0.8\linewidth]{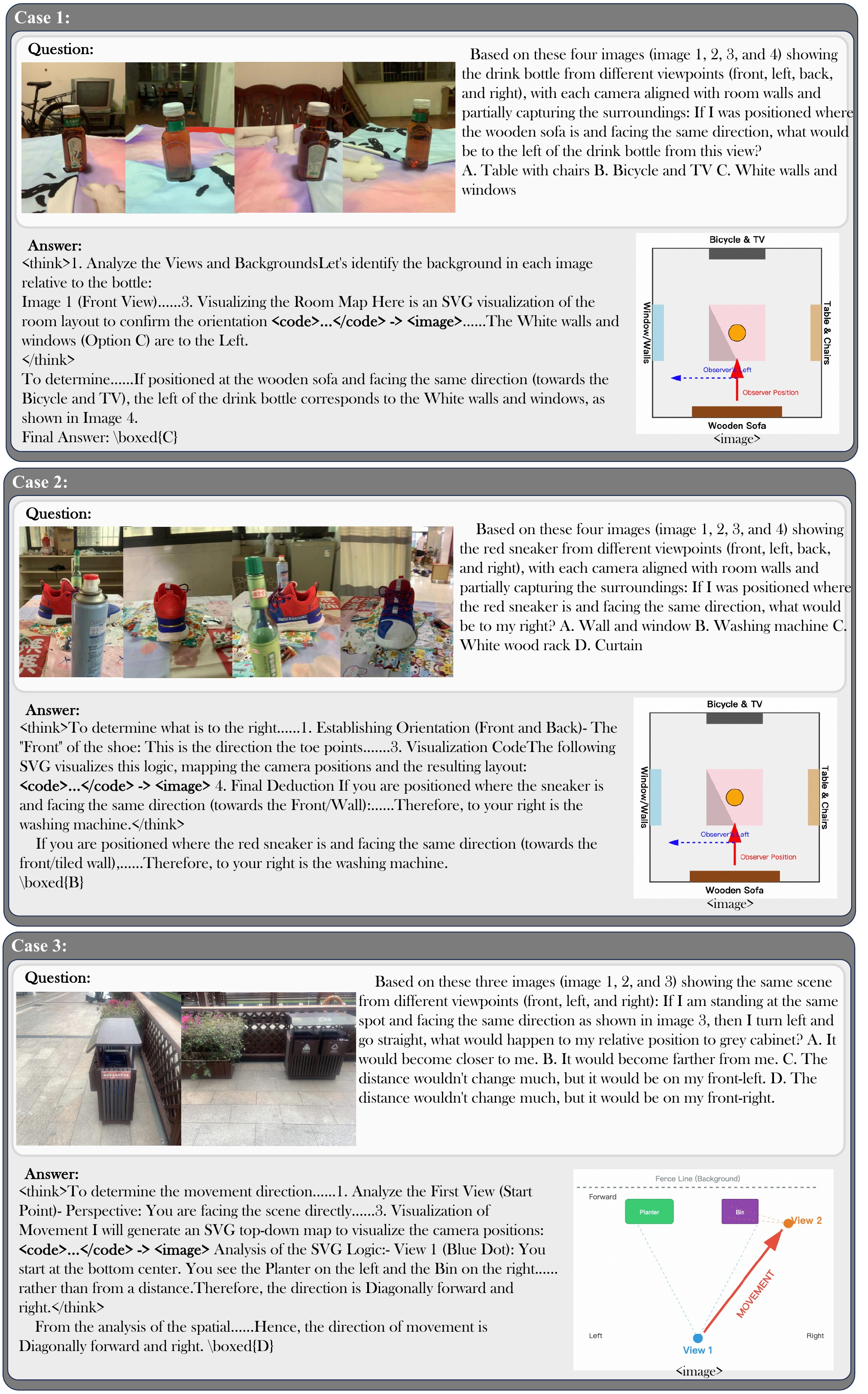}
\vspace{-2mm}
\caption{\textbf{Additional case studies} for \textit{think-with-SVG} reasoning of \textsc{MentalThink} (1/2).}
\label{fig:cases_single_turn1}
\end{figure*}

\begin{figure*}[t]
\centering
\includegraphics[width=0.8\linewidth]{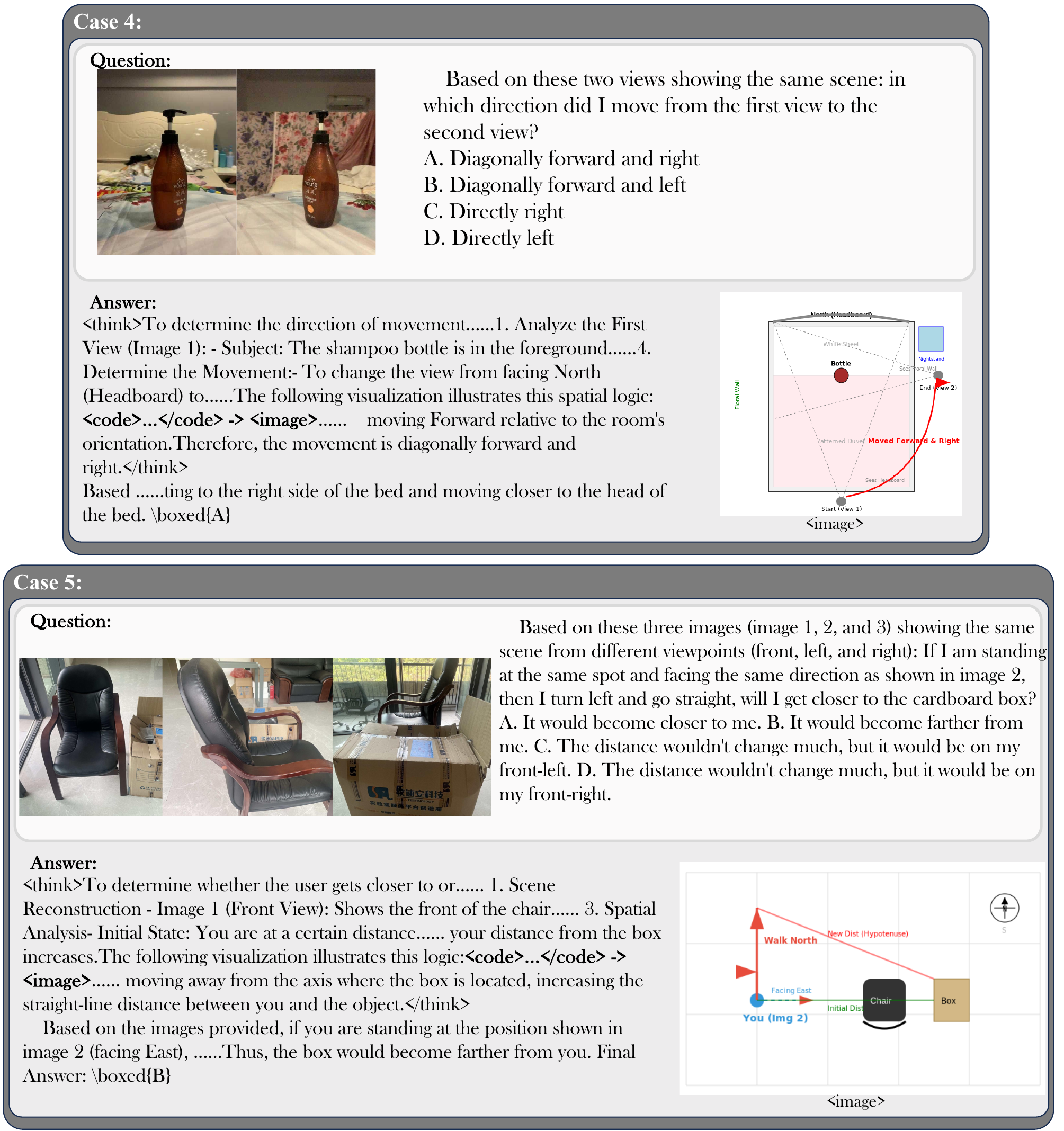}
\vspace{-2mm}
\caption{\textbf{Additional case studies} for \textit{think-with-SVG} reasoning of \textsc{MentalThink} (2/2).}
\label{fig:cases_single_turn2}
\end{figure*}

We present more case studies to illustrate the SVG reconstruction capability (Figure~\ref{fig:cases_reconstruction}) and the \textit{think-with-SVG} capability (Figures~\ref{fig:cases_single_turn1} and~\ref{fig:cases_single_turn2}) of \textsc{MentalThink}.

\end{document}